\documentclass[final,3p,times]{elsarticle}
\pdfoutput=1

\usepackage{amssymb}

\graphicspath{ {./} }
\usepackage[colorlinks, linkcolor=red, anchorcolor=blue, citecolor=green]{hyperref}
\usepackage{float}
\usepackage{makecell}
\usepackage{longtable}
\usepackage{mathtools}
\usepackage[ruled,vlined]{algorithm2e}
\usepackage{booktabs}
\usepackage{multirow}
\usepackage[normalem]{ulem}
\useunder{\uline}{\ul}{}
\renewcommand{\labelitemii}{\raisebox{.25\height}{\fontsize{4pt}{4pt}$\blacksquare$}}
\usepackage{pdflscape}
\usepackage{threeparttable}
\usepackage{graphicx}

\journal{Information Sciences}

\begin{document}

\begin{frontmatter}



\title{Managing dataset shift by adversarial validation for credit scoring}

\author[label1]{Hongyi Qian}
\ead{qianhongyi@buaa.edu.cn}

\author[label2]{Baohui Wang}
\ead{wangbh@buaa.edu.cn}

\author[label2]{Ping Ma}
\ead{maping@buaa.edu.cn}

\author[label3]{Lei Peng}
\ead{penglei@chamc.com.cn}

\author[label3]{Songfeng Gao}
\ead{gaosongfeng@chamc.com.cn}

\author[label2,label1]{You Song\corref{cor1}}
\ead{songyou@buaa.edu.cn}

\cortext[cor1]{Corresponding author.}
\address[label1]{School of Computer Science and Engineering, Beihang University, Beijing 100191, PR China}
\address[label2]{School of Software, Beihang University, Beijing 100191, PR China}
\address[label3]{HuaRong RongTong (Beijing) Technology Co., Ltd, Beijing 100033, PR China}

\begin{abstract}
Dataset shift is common in credit scoring scenarios, and the inconsistency between the distribution of training data and the data that actually needs to be predicted is likely to cause poor model performance. However, most of the current studies do not take this into account, and they directly mix data from different time periods when training the models. This brings about two problems. Firstly, there is a risk of data leakage, i.e., using future data to predict the past. This can result in inflated results in offline validation, but unsatisfactory results in practical applications. Secondly, the macroeconomic environment and risk control strategies are likely to be different in different time periods, and the behavior patterns of borrowers may also change. The model trained with past data may not be applicable to the recent stage. Therefore, we propose a method based on adversarial validation to alleviate the dataset shift problem in credit scoring scenarios. In this method, partial training set samples with the closest distribution to the predicted data are selected for cross-validation by adversarial validation to ensure the generalization performance of the trained model on the predicted samples. In addition, through a simple splicing method, samples in the training data that are inconsistent with the test data distribution are also involved in the training process of cross-validation, which makes full use of all the data and further improves the model performance. To verify the effectiveness of the proposed method, comparative experiments with several other data split methods are conducted with the data provided by Lending Club. The experimental results demonstrate the importance of dataset shift in the field of credit scoring and the superiority of the proposed method.
\end{abstract}



\end{frontmatter}


\section{Introduction}
\label{Introduction}
With the rapid development of internet finance in recent years, users can simply use online platforms to complete peer-to-peer transactions without going through the complex approval process of banks. As the core business of online financial institutions, credit loans not only bring huge profits to the platform but also cause many credit risk problems. How to effectively evaluate the borrowers' solvency based on multi-dimensional information and reduce default risk has become an important research area in the academic and business community \citep{doi:10.1126/science.1200138, Maldonado2020}.

At this stage, with the continuous development of intelligent machine learning methods, credit scoring models have made a series of progress in balanced sampling method \citep{Moscato2021, Niu2020}, feature selection \citep{Lopez2019, Kozodoi2019}, and ensemble model \citep{Marques2012, Xia2020}. These advancements have allowed credit scoring to reach new heights of accuracy, but the vast majority of them still use traditional cross-validation schemes for data segmentation \citep{Song2020, Xia2018, Xiao2021}. The whole dataset is randomly split without considering the dataset shift problem.

\begin{figure}[H]
    \centering
    \includegraphics[width=12cm]{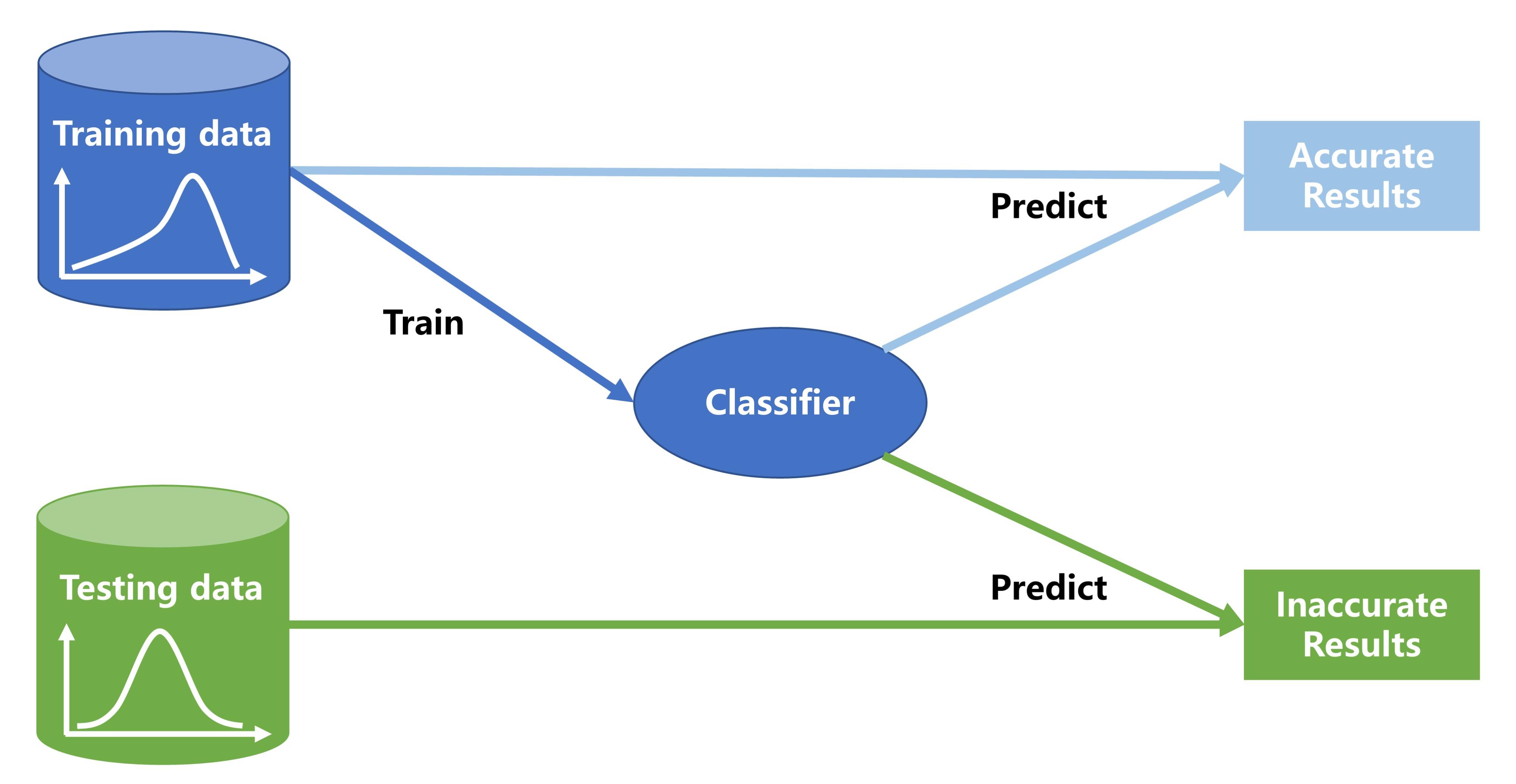}
    \caption{Classifier performance loss caused by different data distribution.}
    \label{fig: different distribution}
\end{figure}

Dataset shift \citep{10.5555/1462129} is an important topic in machine learning. As shown in Fig \ref{fig: different distribution}. it refers to the scenario where the joint distribution of inputs and outputs is inconsistent during the training and testing phases. This inconsistency is usually caused by sample selection bias, which can lead to a loss in the generalization performance of the model on new data. Applications such as demand prediction \citep{10.1145/2523813}, customer profiling for marketing \citep{10.1145/2523813}, and recommender systems \citep{10.1145/2523813, DBLP:journals/cim/DitzlerRAP15} are susceptible to dataset shift. This phenomenon is particularly evident in the non-stationary environments such as credit scoring \citep{doi:10.1057/jors.2009.69}, where changes in the macroeconomic environment and risk control strategies can invalidate models trained with past data.

However, there are few studies on dataset shift in the field of credit scoring. For example, Bravo et al. \citep{Bravo2015} proposed a model-dependent backtesting strategy designed to identify significant changes in the covariates, relating a confidence zone of the change to a maximal deviance measure obtained from the coefficients of the model. This rule-based statistical method has poor generalization performance on different datasets. Maldonado et al. \citep{Maldonado2021} proposed an algorithmic-level machine learning solution, using novel fuzzy support vector machine (FSVM) strategy, in which the traditional hinge loss function is redefined to account for dataset shift. However, to our best knowledge, no data-level machine learning solution has been proposed to solve the dataset shift problem in the field of credit scoring.

The main reason why the dataset shift problem has not been highlighted in the credit scoring field is that the major credit scoring public datasets in the past do not provide the timestamp information of the samples. Such as German \citep{Dua:2019}, Australian \citep{Dua:2019}, Taiwan \citep{Dua:2019, DBLP:journals/eswa/YehL09a}, Japan \citep{Dua:2019} in the UCI repository\footnote{\url{https://archive.ics.uci.edu/ml/index.php}} and PAKDD\footnote{\url{https://pakdd.org/archive/pakdd2009/front/show/competition.html}}, Give Me Some Credit\footnote{\url{https://www.kaggle.com/c/GiveMeSomeCredit}}, Home Credit Default Risk\footnote{\url{https://www.kaggle.com/c/home-credit-default-risk/data}} provided in the data mining competitions. As a result, researchers are unable to construct the training and testing sets in chronological order.

However, that can be changed with the release of the Lending Club\footnote{\url{https://www.lendingclub.com/}} dataset. Lending Club is a US peer-to-peer lending company, headquartered in San Francisco, California. It was the first peer-to-peer lending institution to register its offerings as securities with the Securities and Exchange Commission (SEC), and to offer loan trading on a secondary market. Lending Club has grown into the world's largest peer-to-peer lending platform, and it provided a large number of real credit data for practitioners and scholars to study. The provided data have specific timestamp information that allows researchers to easily study the effect of dataset shift on the model's performance on the latest data.

The goal of this paper is to propose a data-level machine learning solution to deal with the problem of dataset shift in credit scoring scenarios. The proposed methods are based on adversarial validation to ensure the generalization performance of the trained model on the predicted samples. Specifically, for the best solution, cross-validation is performed by selecting partial samples in the training set that are closer to the distribution of the predicted data through adversarial validation. In addition, the remaining training samples that are not consistent with the distribution of the test data are also involved in the cross-validation training process, but not in the validation, which makes full use of all the data and further improves the model performance. To sum up, the main contributions of this paper are as follows:

\renewcommand{\theenumi}{\roman{enumi}}
\begin{enumerate}
    \item This paper is the first to consider the dataset shift problem on Lending Club data, which is also the first solution based on a data-level machine learning approach to address the dataset shift in the credit scoring field. Dataset shift is an important topic in machine learning, but there is little research related to it in the credit scoring field \citep{Bravo2015, Maldonado2021}. This paper recommends paying more attention to the impact of data distribution on the model effectiveness, rather than just minimizing the classification error.
    \item The method used to solve the dataset shift problem in this paper is based on adversarial validation. Uber researchers \citep{pan2020adversarial} have previously proposed a method that uses adversarial validation to filter features to deal with dataset shift. However, there is a trade-off for this method between the improvement of generalization performance and losing information. On the contrary, the method proposed in this paper based on adversarial validation can make full use of all data to improve the model generalization performance.
    \item Experiments on Lending Club data showed that the proposed method in this paper achieves the best results compared to the existing methods that commonly use cross-validation or timeline filtering to partition data.
\end{enumerate}

The rest of this paper is organized as follows. \hyperref[Dataset shift]{Section 2} presents some theoretical background of dataset shift. \hyperref[Methodology]{Section 3} details the adversarial validation based method to help balance the training and testing sets. \hyperref[Experimental study]{Section 4} shows the design details and results of the experiments and discusses them. \hyperref[Conclusion]{Section 5} gives the conclusion and illustrates the direction for future research.

\section{Dataset shift}
\label{Dataset shift}

\subsection{Definition of dataset shift}
The term \textit{dataset shift} was first introduced by J. Quionero-Candela et al \citep{10.5555/1462129}. In other studies, it has also been called the \textit{concept shift} \citep{DBLP:journals/ml/WidmerK96}, \textit{changes of classification} \citep{DBLP:conf/sdm/WangZFY03}, \textit{fracture points} \cite{DBLP:journals/kais/CieslakC09} or \textit{contrast mining in classification learning} \citep{DBLP:journals/dke/YangWZ08}. Such inconsistent terminology confounds the discussion of this important problem. In this paper, the term \textit{dataset shift} is represented for the situation where the data used to train the classifier and the environment where the classifier is deployed do not follow the same distribution, which means $P_{train}(y, x) \neq P_{test}(y, x)$.

\subsection{Types of dataset shift}
A classification problem consists of three parts, namely a set of features or covariates $x$, a target variable $y$, and joint distribution $P(y, x)$. There are two kinds of classification problems according to \citep{Moreno-Torres2012}:

\begin{itemize}
    \item $X \rightarrow Y$ problems, where the class label is causally determined by the values of the covariates. For example, user behavior in credit scoring, represented by the covariable space $X$, determines the class label $Y$: good or bad users.
    \item $Y \rightarrow X$ problems, where the class label causally determines the values of the covariates. Medical diagnostics typically fall into this category, where the disease, which is modeled as the class label $Y$, determines the symptoms, represented as covariates $X$ in the machine learning task.
\end{itemize}

There are three different types of shift for the above two kinds of problems, depending on which probabilities change or not:
\begin{itemize}
    \item \textbf{Covariate shift} represents the situation where training and testing data distribution may differ arbitrarily, but there is only one unknown target conditional class distribution. In other words, it appears only in $X \rightarrow Y$ problems, and is mathematically defined as the case where $P_{train }(y \mid x)=P_{test }(y \mid x)$ and $P_{train }(x) \neq P_{test }(x)$.
    \item \textbf{Prior probability shift} is the reverse case of covariate shift. It appears only in $Y \rightarrow X$ problems, and is defined as the case where $P_{train }(x \mid y)=P_{test }(x \mid y)$ and $P_{train }(y) \neq P_{test }(y)$.
    \item \textbf{Concept shift} happens when the relationship between the input and class variables changes, which is defined as
    \begin{itemize}
        \item[\labelitemii] $P_{train }(y \mid x) \neq P_{test }(y \mid x)$ and $P_{train }(x) = P_{test }(x)$ in $X \rightarrow Y$ problems.
        \item[\labelitemii] $P_{train }(x \mid y) \neq P_{test }(x \mid y)$ and $P_{train }(y) = P_{test }(y)$ in $Y \rightarrow X$ problems.
    \end{itemize}
\end{itemize}

Both the covariate shift and the concept shift can occur in the credit scoring scenario. One example of the covariate shift is that as the economy grows and per capita wages rise, the old model of using a person's income to determine his or her credit rating may slowly fail. For concept Shift, a common example is a sudden change in the macroeconomic environment or loan policy. The risk level may change after the loan interest rate change even for the same lender.

\subsection{Causes of dataset shift}
There are many possible reasons for dataset shift, the two most important of which are as follows:

\textbf{Reason 1. Sample Selection Bias} is a systematic defect in the data collection or labeling process, where the training set is obtained by a biased method and this non-uniform selection will cause the training set to fail to represent the real sample space. For example, in social science research, students at the researcher's university or former research participants are more likely to be surveyed than other populations. These ``easy'' groups may be overrepresented in the training samples, while ``difficult'' groups (e.g., prisoners) may be underrepresented or completely excluded.

Joaquin et al. \citep{QuioneroCandela2009WhenTA} give a mathematical definition of sample selection bias:

\begin{itemize}
    \item $P_{train}=P(s=1 \mid y,x)P(x)$ and $P_{test} = P(y \mid x)P(x)$ in $X \rightarrow Y$ problems.
    \item $P_{train}=P(s=1 \mid x,y)P(y)$ and $P_{test} = P(x \mid y)P(y)$ in $Y \rightarrow X$ problems.
\end{itemize}
where $s$ is a binary selection variable that decides whether an instance is included in the training samples ($s=1$) or rejected from it ($s=0$).

The problem of operating under sample selection bias has received substantially more attention in other domains than it has in the machine learning community \citep{Moreno-Torres2012}. In the credit scoring literature it goes by the name of reject inference, because potential credit applicants who are rejected under the previous model are not available to train future models \citep{Crook2004DoesRI, Hand1997StatisticalCM}.

\textbf{Reason 2. Non-stationary Environments} is often caused by temporal or spatial changes, and is very common in real-world applications. Depending on the classification problem's type, non-stationary environments can lead to different kinds of shift:

\begin{itemize}
    \item In $X \rightarrow Y$ problems, a non-stationary environment could create changes in either $P(x)$ or $P(y \mid x)$, generating covariate shift or concept shift, respectively.
    \item In $Y \rightarrow X$ problems, it could generate prior probability shift with a change in $P(y)$ or concept shift with a change in  $P(x \mid y)$.
\end{itemize}

Non-stationary environments often appears in adversarial classification problems such as network intrusion detection \citep{Kolcz2009FeatureWF}, spam detection \citep{10.1145/1014052.1014066, Barreno2010TheSO} and fraud detection \citep{Fawcett2004AdaptiveFD, 10.1145/347090.347102}. The presence of an adversary trying to bypass the existing classifier introduces any possible dataset shift, and the bias can change dynamically. This kind of problem is getting more and more attention in the machine learning community \citep{Biggio2010MultipleCS, Laskov2010MachineLI}.

\subsection{Solutions for dataset shift}

A common approach to cope with the dataset shift problem in production systems is monitoring and redeploying, i.e., continuous monitoring the model performance and retraining the model with new data when it degrades but does not completely fail \citep{10.1145/2523813}. Throughout the process, the model goes through a cycle of built, deployed, deprecated, and rebuilt, and this cyclical strategy is important to maintain an accurate and stable system. In the field of credit scoring, Kolmogorov-Smirnov (KS) and Population Stability Index (PSI) are often used as indicators to monitor model performance \citep{Siddiqi2005CreditRS}. Although this approach can reduce the impact of dataset shift issues to some extent, it still encounters many challenges in practical use. For example, model predictions usually take some time to show their effects, and there is no immediate observable feedback. To solve this problem, some scholars have proposed intelligent machine learning methods to directly deal with dataset shift, which are divided into two main categories, algorithmic-level and data-level solutions.

Algorithmic-level solutions propose models that are robust in the presence of dataset shift. There are many types of algorithmic-level solutions, of which the Bayesian model is a common one, but they are mostly designed for regression rather than classification problems, such as the stock price \citep{Dangl2012} or temperature prediction \citep{Storvik2002139}. In the scenario of online learning or data stream mining, many algorithmic-level solutions have also emerged. For example, Kolter et al. \citep{Kolter2007} proposed dynamic weighted majority (DWM) for concept drift that dynamically creates and removes weighted experts in response to changes in performance. In addition to that, Mello et al. \citep{Mello2020} combine the twin support vector machine with a fuzzy membership function (FBTWSVM) to deal with large datasets and learn from data streams. In the field of credit scoring, Maldonado et al. \citep{Maldonado2021} proposed a general version of hinge loss function applying aggregation operators to deal with dataset shift via fuzzy logic.

For data-level solutions, a common approach is the forgetting mechanism \citep{10.1145/2523813, DBLP:journals/ml/WidmerK96}, which removes outdated samples by sliding time windows. There is a tradeoff between reactivity and robustness to noise of the system, and the more abrupt forgetting is, the faster the reactivity, but also the higher the risk is of capturing noise. Another common data-level solution is the novel variant of cross-validation \citep{Lopez2014, Moreno-Torres2012a, Sugiyama2007}, which has the advantage that they do not require timestamps because they are designed to evaluate changes in the data distribution.

\section{Methodology}
\label{Methodology}

\subsection{Adversarial validation}

Adversarial validation is a method to detect dataset shift, which requires training a binary classifier and judging whether the sample is from the training set or the testing set. If the performance of the classifier is close to the result of random guess, it means that it is difficult to distinguish whether a sample is from the training set or the testing set, that is, the distribution of both is relatively consistent. On the contrary, if the classifier performance is much better than the result of random guess, it indicates that the sample distribution of the training set and testing set is inconsistent.

It should be noted that the \textit{adversarial validation} mentioned in this article is not the same as the \textit{adversarial training} introduced in Generative Adversarial Networks (GAN) \citep{Goodfellow2014GenerativeAN}. GAN's framework corresponds to a minimax two-player game, it simultaneously trains two models: a generative model $G$ that captures the data distribution, and a discriminative model $D$ that estimates the probability that a sample came from the training data rather than $G$. GAN is becoming more and more popular in the field of content generation \citep{Zhu2017UnpairedIT, 9157662}. In the field of credit scoring, GAN has been used to solve the sample imbalance problem \citep{Engelmann2021}.

Specifically, the process of adversarial validation can be divided into three steps:

\renewcommand{\theenumi}{\roman{enumi}}
\begin{enumerate}
    \item For the original dataset $\left\{train\_X, train\_y, val\_X, val\_y\right\}$, remove the old label column $\left\{train\_y,val\_y\right\}$, and add a new label column $\left\{train\_y_{s},val\_y_{s}\right\}$ that marks the source of the data, labeling the samples in the training set as 0 $(i.e.\ train\_y_{s}=0)$ and the samples in the testing set as 1 $(i.e.\ train\_y_{s}=1)$.
    \item Train the classifier on the dataset $\left\{train\_X, train\_y_{s}, val\_X, val\_y_{s}\right\}$ with the newly labeled column. The output of the classifier is the probability that the sample belongs to the testing set. In this paper, 5-fold cross-validation is used.
    \item Observe the results of the classifier. If the model performance is poor (AUC score is close to 0.5), it indicates that the classifier cannot distinguish whether the samples come from the training set or the testing set. It can be judged that the distribution of the training set and testing set in the original data is consistent. On the contrary, if the model performance is high (AUC score is close to 1), it indicates that the classifier can easily distinguish sample sources. It can be determined that the training set and the testing set are very different in data distribution.
\end{enumerate}

In addition to detecting inconsistent data distribution, the results of adversarial validation can further help balance training and testing sets, and improve the model performance on testing set. This appears in some data competition, but there is still relatively little published research. Researchers at Uber \citep{pan2020adversarial} proposed an adversarial validation based approach that addresses the issue of concept drift that commonly exists in large-scale user targeting automation systems, where the distribution of complex user features keeps evolving. They use the feature importance obtained from adversarial validation to filter the most inconsistently distributed features sequentially. However, there is a trade-off for this method between the improvement of generalization performance and losing information by dropping features from the model. The proposed method in this paper can improve the generalization performance of the model to the testing set without losing information.

\subsection{The adversarial validation based method to deal with dataset shift}

The method proposed based on adversarial validation in this paper can not only judge whether the dataset distribution is consistent, but also further balance the training and testing sets. Specifically, gradient boosting decision tree (GBDT) \citep{Friedman2001} is used as the classifier for both the adversarial validation and the credit scoring phases, which is a boosting model that continuously reduces the residuals during the training process. It was chosen because the GBDT-based methods have been proven to be very effective in recent research in the field of credit scoring \citep{He2018, Liu2021}. Apart from that, GBDT has a very efficient modern library\footnote{\url{https://github.com/microsoft/LightGBM}}, LightGBM\citep{Ke2017}, which has won many data competitions and will be used in this paper to build the model.

There are many ways to use the results of adversarial validation, and a total of three schemes are proposed in this paper.

\textbf{Method 1.} Use the adversarial validation results as sample weights added to the training process.

The adversarial validation probability result of the training set sample represents its similarity to the testing set, which can be used to determine how much to involve in the actual modeling process. If a sample is more closely distributed to the testing set data, it can be given a higher weight during training. Conversely, lower weights are assigned. Modern GBDT libraries can specify the weight of each sample during training, which is convenient to control the contribution of each sample. Specifically, set the ``weight" parameter in LightGBM Dataset API to change the weight of each instance.

\textbf{Method 2.} Use only the data with the top-ranked adversarial validation results for 5-fold cross-validation.

Apart from reducing the weight of samples inconsistent with the distribution of the testing set, they can also be removed from the training process. In particular, the training data can be divided into two parts by screening the results of adversarial validation probability according to a certain threshold value. The samples that are more consistent with the testing set distribution are called $data\_X_{a}=\left\{train\_X_{a}, val\_X_{a}\right\}$, and the remaining samples are called $data\_X_{b}$. $P_{data\_X_{a}} \approx P_{test\_X} \neq P_{data\_X_{b}}$, and only $data\_X_{a}$ is reserved for 5 fold cross-validation.

As a result, model evaluation metrics on the validation data should have similar results on the testing data, which means that if the model works well on the validation data, it should also work well on the testing data. Specifically, The LightGBM parameter ``early\_stopping\_round" is set during training, and the model stops training when the AUC of the validation set $val\_X_{a}$ still doesn't grow in a certain number of iterations. The model with optimal results is retained and used to predict the testing set.

\textbf{Method 3.} All data are used for training, and only the data with the top-ranked adversarial validation results are used for validation.

Although Method 2 alleviates the problem of inconsistent data distribution between training and testing sets, it has defects in data utilization. Only $data\_X_{a}$ data was used in the whole training process, and $data\_X_{b}$ data was wasted. To solve this problem, $data\_X_{b}$ is added to the training data of each fold in the process of 5-fold cross-validation to assist training, but it does not participate in the validation. This not only maintains the consistency of validation and testing results, but also makes full use of all data.

\section{Experimental study}
\label{Experimental study}

\subsection{Data collection}
The dataset used in this paper comes from Lending Club, whose timestamp information helps divide training and testing sets in strict chronological order. The data includes samples from 2018M1\footnote{The representation of time in this paper consists of two parts, M is preceded by the year and followed by the month, for example, 2018M1 represents January 2018.} to 2020M9 over a period of 33 months. The original dataset contains 1,160,066 samples with 151 features and ``loan\_status" as the target variable. As shown in Table \ref{table: Dataset characterization}, ``loan\_status'' has 8 states, ``Current'', ``Fully Paid'', ``Charged Off'', ``Late (31-120 days)'', ``In Grace Period'', ``Issued'', ``Late (16-30 days)'', and ``Default''. Referring to the practice of previous papers \citep{Moscato2021, Namvar2018}, only the samples with ``Charged off'' and ``Fully Paid'' status are taken as positive and negative samples respectively, all loans with other status have been filtered out as their final status are unknown. This results in an unbalanced dataset containing 276,685 samples with a positive sample ratio of 21.93\%.

\begin{table}[H]\footnotesize
    \centering
    \caption{Dataset characterization.}
    \label{table: Dataset characterization}
    \begin{tabular}{ll}
    \hline
    Loan status & Amount \\ \hline
    Current & 859,320 \\
    Fully Paid & 216,019 \\
    Charged Off & 60,666 \\
    Late (31-120 days) & 12,283 \\
    In Grace Period & 7,476 \\
    Issue & 2,062 \\
    Late (16-30 days) & 2,056 \\
    Default & 184 \\
    \textbf{Total} & 1,160,066 \\ \hline
    \end{tabular}
\end{table}

\begin{figure}[H]
    \centering
    \includegraphics[width=\textwidth]{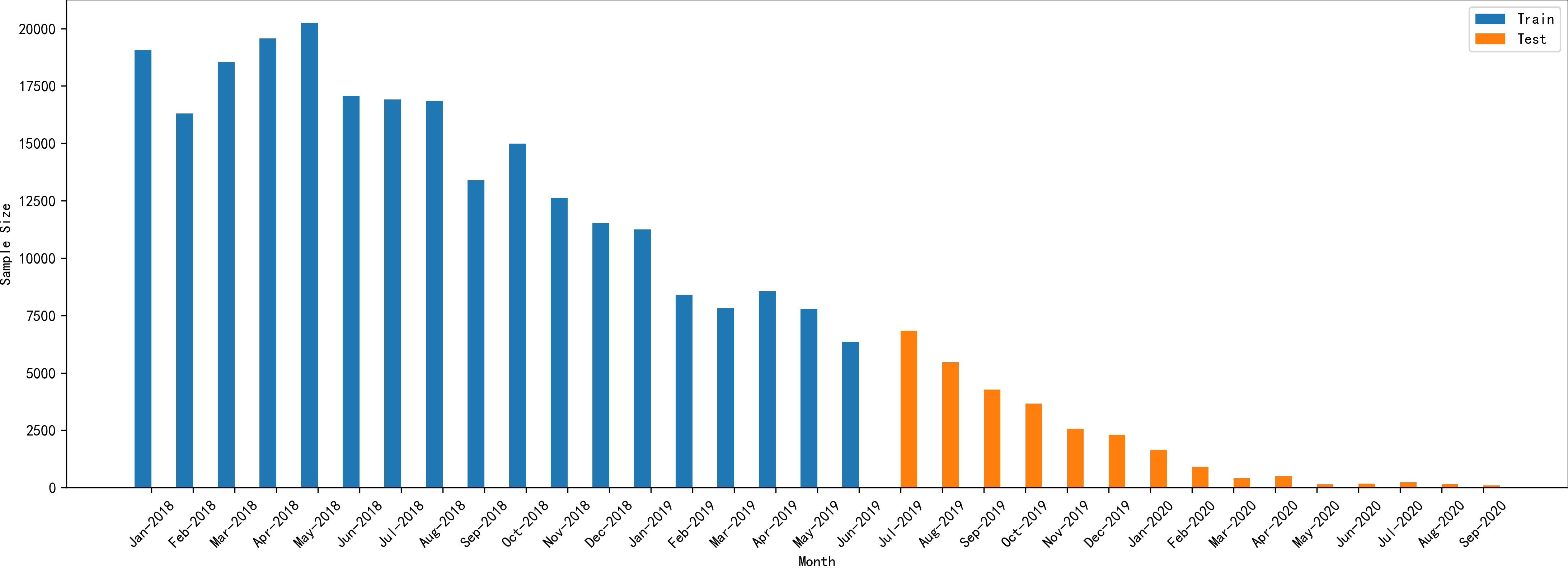}
    \caption{Sample size distribution by month.}
    \label{fig: Sample size distribution by month}
\end{figure}

The number of samples in each month is shown in Fig \ref{fig: Sample size distribution by month}. In chronological order, the data of 18 months from 2018M1 to 2019M6 are taken as the training set, which contains 247,276 samples in total. The data from 2019M7 to 2020M9 were taken as the testing set, including 29,409 samples.

Many features in the original data have a high proportion of missing values, and some of the remaining variables are unavailable to an investor before deciding to fund the loan. As a result, including the target variable ``loan\_status'', 25 variables are actually used, and each field description is shown in table \ref{table: Indicator descriptions}.

The specific processing methods for some variables are as follows:
\begin{itemize}
    \item The original ``emp\_length'' variable contains $<$1 year, 1 year ... 9 years, 10+ years for a total of 11 category variables, which are transformed into integer variables from 0 to 10 for better usage;
    \item The original FICO score provides two values, ``fico\_score\_low'' and ``fico\_score\_high'', and the difference between the two is always 4 points, which leads to data redundancy and is replaced by the average of the two;
    \item Log transformation is performed on numerical variables with power-law distribution, including ``annual\_inc'' and ``revol\_bal'';
    \item For category variables, including ``sub\_grade'', ``home\_ownership'', ``verification\_status'', ``initial\_list\_status'', ``purpose'', ``addr\_state'', ``application\_type'', we employ the LightGBM built-in support.
\end{itemize}

\begin{table}[H]\footnotesize
    \centering
    \caption{Indicator descriptions.}
    \label{table: Indicator descriptions}
    \begin{tabular}{ll}
    \hline
    LoanStatNew & Description \\ \hline
    addr\_state & The state provided by the borrower in the loan application. \\
    annual\_inc & The self-reported annual income provided by the borrower during registration. \\
    application\_type & Indicates whether the loan is an individual application or a joint application with two co-borrowers. \\
    dti & Borrower’s total monthly debt payments divided by the borrower’s self-reported monthly income. \\
    earliest\_cr\_line & The month the borrower's earliest reported credit line was opened. \\
    emp\_length & Employment length in years. \\
    fico\_range\_high & The upper boundary range the borrower’s FICO at loan origination belongs to. \\
    fico\_range\_low & The lower boundary range the borrower’s FICO at loan origination belongs to. \\
    home\_ownership & The home ownership status provided by the borrower during registration. \\
    initial\_list\_status & The initial listing status of the loan. \\
    installment & The monthly payment owed by the borrower if the loan originates. \\
    int\_rate & Interest Rate on the loan. \\
    loan\_amnt & The listed amount of the loan applied for by the borrower. \\
    loan\_status & Current status of the loan. \\
    mort\_acc & Number of mortgage accounts. \\
    open\_acc & The number of open credit lines in the borrower's credit file. \\
    pub\_rec & Number of derogatory public records. \\
    pub\_rec\_bankruptcies & Number of public record bankruptcies. \\
    purpose & A category provided by the borrower for the loan request. \\
    revol\_bal & Total credit revolving balance. \\
    revol\_util & Revolving line utilization rate, or the amount of credit relative to all available revolving credit. \\
    sub\_grade & LC assigned loan subgrade. \\
    term & The number of payments on the loan. Values are in months and can be either 36 or 60. \\
    total\_acc & The total number of credit lines currently in the borrower's credit file. \\
    verification\_status & Indicates if income was verified by LC, not verified, or if the income source was verified. \\ \hline
    \end{tabular}
\end{table}

After the above processing, the actual number of variables input to the model is 23. Table \ref{table: Indicators statistics} shows the statistical characteristics of the numerical variables.

\begin{table}[H]\footnotesize
    \centering
    \caption{Indicators statistics.}
    \label{table: Indicators statistics}
    \begin{tabular}{lllllllll}
    \hline
    variable name & count & mean & std & min & 25\% & 50\% & 75\% & max \\ \hline
    loan\_amnt & 276685 & 15202.28175 & 10022.38 & 1000 & 7500 & 12000 & 20000 & 40000 \\
    term & 276685 & 42.295072 & 10.55719 & 36 & 36 & 36 & 60 & 60 \\
    int\_rate & 276685 & 13.27946 & 5.383279 & 5.31 & 8.81 & 12.4 & 16.46 & 30.99 \\
    installment & 276685 & 454.716117 & 291.7174 & 28.77 & 238.02 & 372.88 & 619.47 & 1676.23 \\
    emp\_length & 252697 & 5.729265 & 3.757685 & 0 & 2 & 5 & 10 & 10 \\
    loan\_status & 276685 & 0.21926 & 0.413746 & 0 & 0 & 0 & 0 & 1 \\
    dti & 275967 & 19.339198 & 21.08527 & 0 & 11.08 & 17.37 & 24.68 & 999 \\
    earliest\_cr\_line & 276685 & 2002.429882 & 7.689656 & 1944 & 1999 & 2004 & 2007 & 2017 \\
    open\_acc & 276685 & 11.509359 & 5.920323 & 0 & 7 & 10 & 14 & 86 \\
    pub\_rec & 276685 & 0.145812 & 0.39563 & 0 & 0 & 0 & 0 & 52 \\
    revol\_util & 276289 & 40.718864 & 25.19198 & 0 & 20.4 & 38 & 59 & 146.3 \\
    total\_acc & 276685 & 23.802024 & 12.615 & 2 & 15 & 22 & 31 & 148 \\
    mort\_acc & 276685 & 1.456483 & 1.784587 & 0 & 0 & 1 & 2 & 24 \\
    pub\_rec\_bankruptcies & 276685 & 0.136889 & 0.350828 & 0 & 0 & 0 & 0 & 7 \\
    log\_annual\_inc & 276685 & 4.816797 & 0.35534 & 0 & 4.676703 & 4.832515 & 4.986776 & 6.968483 \\
    fico\_score & 276685 & 709.598766 & 36.96096 & 662 & 682 & 702 & 732 & 847.5 \\
    log\_revol\_bal & 276685 & 3.883877 & 0.687627 & 0 & 3.665769 & 3.990605 & 4.26257 & 6.167597 \\ \hline
    \end{tabular}
\end{table}

\subsection{Parameter set-up}

The same settings are used for LightGBM hyperparameters in both the adversarial validation and credit scoring phases. Specifically, ``num\_boost\_round'' is set to 50000, which is a relatively large value. However, by setting the ``early\_stopping\_rounds'' parameter, the model will stop training if the validation data's AUC doesn't improve in the last 200 rounds. It not only ensures sufficient training, but also prevents over-fitting. More detailed hyperparameter settings are shown in Table \ref{table: Parameters of LightGBM}.

\begin{table}[H]\footnotesize
    \centering
    \caption{Parameters of LightGBM.}
    \label{table: Parameters of LightGBM}
    \begin{tabular}{ll}
    \hline
    Parameters & Values \\ \hline
    num\_boost\_round & 50000 \\
    early\_stopping\_rounds & 200 \\
    learning\_rate & 0.1 \\
    max\_depth & 4 \\
    num\_leaves & 8 \\
    colsample\_bytree & 0.8 \\
    subsample & 0.8 \\
    subsample\_freq & 3 \\
    the others & default \\ \hline
    \end{tabular}
\end{table}

\subsection{Experiment set-up}
\begin{figure}[H]
    \centering
    \includegraphics[width=\textwidth]{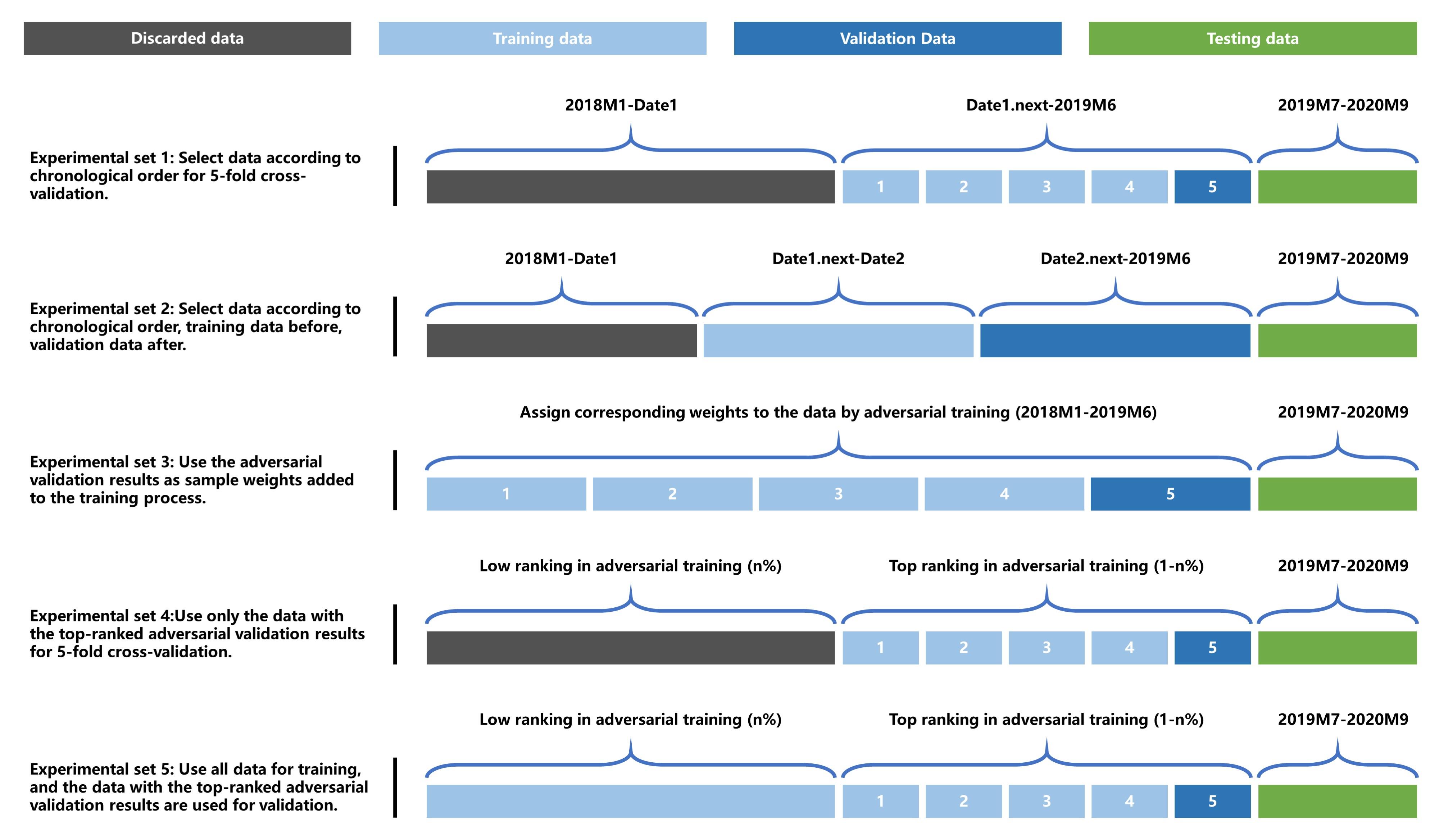}
    \caption{Experimental setup.}
    \label{fig: experiment setup}
\end{figure}

As shown in Fig \ref{fig: experiment setup}, in order to demonstrate the effect of adversarial validation, a total of 5 sets of experiments are set up. The testing set data of each experiment are all from 2019M7 to 2020M9, while the training and validation sets are divided in different ways.

\textbf{Experimental Set 1.} Select data according to chronological order for 5-fold cross-validation.

A fixed time point is set and only data after that time point are used for the 5-fold cross-validation. Specifically, the starting month of the cross-validation data was selected from 2018M1 to 2019M6 for a total of 18 experiments. In these experiments, the starting point selection 2018M1, which uses all data for training can be used as a benchmark.

\textbf{Experimental Set 2.} Select data according to chronological order, training data before, validation data after.

Different from the 5 fold cross-validation used in Experimental Set 1, experiment set 2 only used data closer to the testing set for validation. Specifically, there are three choices of data time ranges, which are to use all data, 2018M6 and subsequent data, 2018M12 and subsequent data. These three groups of data will be divided into training and validation data according to the sequence of timeline, for a total of 17+11+5=33 experiments.

\textbf{Experimental Set 3.} Use the adversarial validation results as sample weights added to the training process.

The output probability of the adversarial validation classifier to the samples, i.e., the similarity with the testing set samples, is directly used as the weight in the training process. All data will be used in this one experiment, and no need to be divided by time or quantile.

\textbf{Experimental Set 4.} Use only the data with the top-ranked adversarial validation results for 5-fold cross-validation.

The ranking of the output probability of the sample by the adversarial validation classifier is regarded as the criterion of data partitioning. Data that is more inconsistent with the distribution of the testing set will be discarded, and the remaining data will be subjected to 5-fold cross-validation. Specifically, 0\%, 5\% … 90\%, 95\% of the data were discarded, respectively, for a total of 20 experiments.

\textbf{Experimental Set 5.} All data are used for training, and only the data with the top-ranked adversarial validation results are used for validation.

Although Experimental Set 4 constructs a dataset that is more consistent with the testing set distribution for cross-validation, it brings about the problem of wasting a lot of data. This can also harm the model performance, especially when the amount of discarded data is large. Experimental Set 5 is further optimized based on Experimental Set 4, by adding these discarded data that are inconsistent with the testing set distribution into the cross-validation training data, but does not participate in the validation. This not only addresses the problem of dataset shift, but also makes full use of all data. Similarly, the data are also divided according to the output probability ranking of the samples by the adversarial validation classifier, and the number of experiments is the same as in Experimental Set 4, with a total of 20 experiments.

There is a total of 18+33+1+20+20=92 experiments in 5 sets, comparing a variety of models' performance on the testing set, which is trained on data divided by time or by adversarial validation results.

\subsection{Results and discussion}

\begin{figure}[H]
    \centering
    \includegraphics[width=8cm]{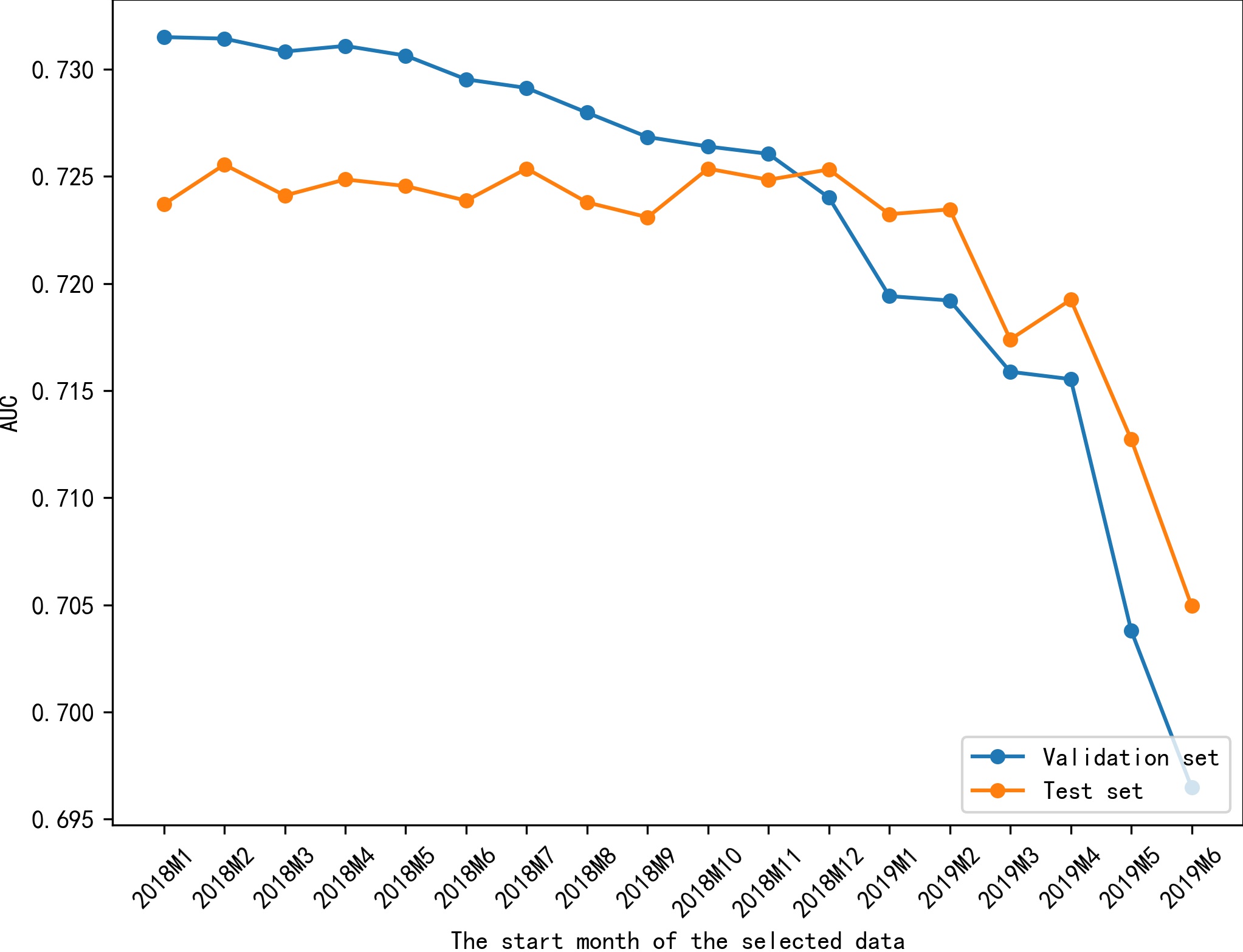}
    \caption{Experimental Set 1 results.}
    \label{fig: experimental set 1}
\end{figure}

\textbf{Analysis of Experimental Set 1.}

Fig \ref{fig: experimental set 1} shows the results of Experimental Set 1. With the increase of the starting month of the selected data, the AUC of the validation set shows a trend of gradual decline, and the decline speed increases with the decrease of the selected data. For the testing set, the AUC fluctuated steadily when the selected data started before 2019M2, and only after that did it start to show a significant decreasing trend. Although adding data with a long distance from the testing set in the training improves the offline validation effect of the model, it does not improve the performance of the test data. This confirms that the problem of dataset shift does exist, and the distribution of the data accumulated in the past is not consistent with the data that needs to be predicted in the recent stage, which leads to the problem of model generalization. The 5-fold cross-validation with all the data could be used as a benchmark, i.e., the starting month of the selected data was set to 2018M1. At this time, the AUC of the testing set was 0.7237. Among all the experiments of Experimental Set 1, the selection of 2018M2 and later data for cross-validation is the best, with the testing set AUC reaching 0.7256.

\textbf{Analysis of Experimental Set 2.}

\begin{figure}[H]
    \centering
    \includegraphics[width=\textwidth]{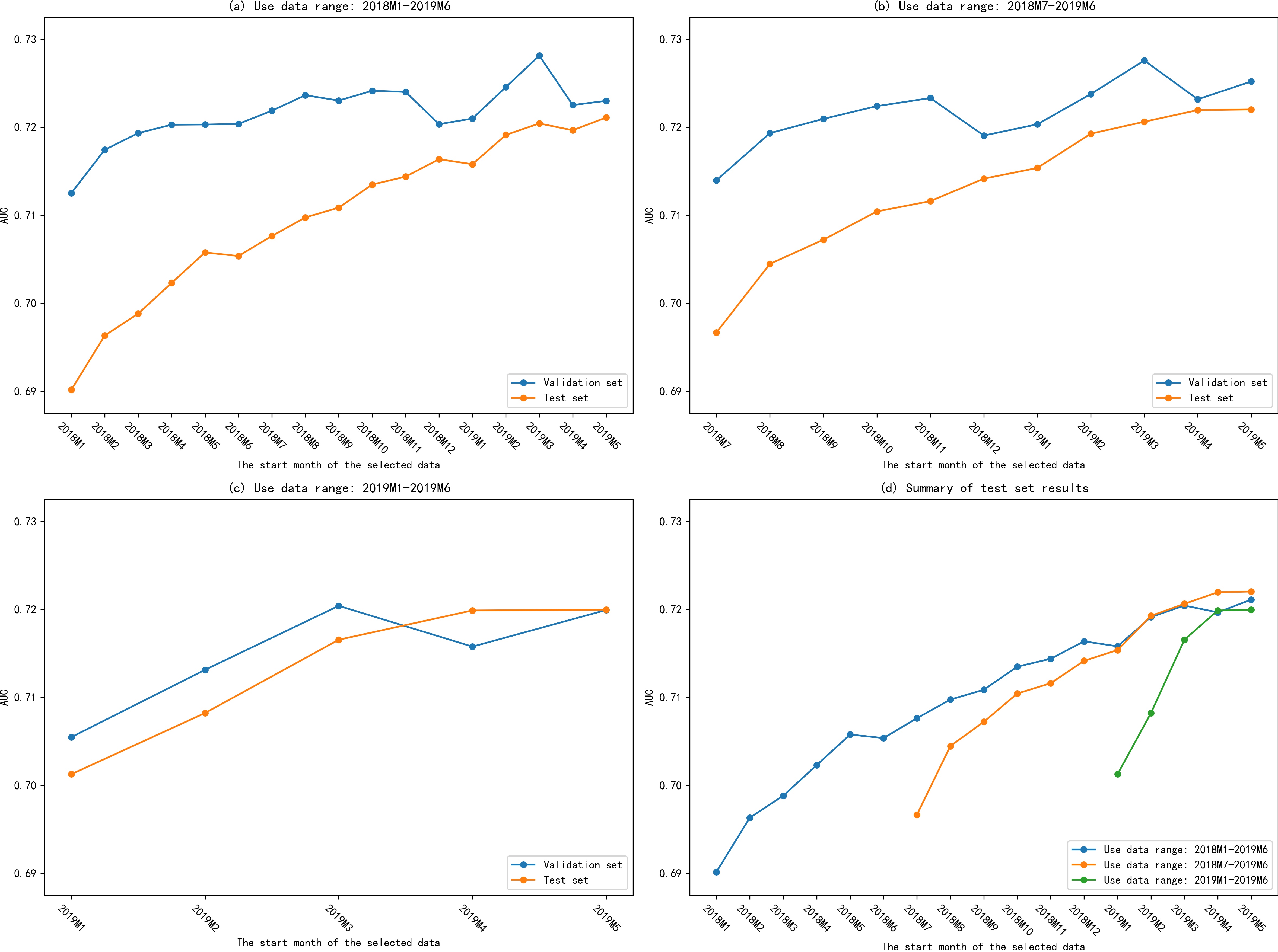}
    \caption{Experimental Set 2 results.}
    \label{fig: experimental set 2}
\end{figure}

Fig \ref{fig: experimental set 2} shows the results of Experimental Set 2. From \ref{fig: experimental set 2} (a) to (c), regardless of the selection range of training validation data starts from 2018M1, 2018M7, or 2019M1, with the gradual increase of data divided into the training set, the AUC of validation set and testing set both show an increasing trend, and the gap between them gradually decreases. This indicates that postponing the time point of splitting the training and validation sets can improve both the performance of the model and the consistency of the validation and testing set results.

Fig \ref{fig: experimental set 2} (d) integrates the testing set AUC results of the three sub-experiments, and the optimal results that can be achieved by all three are relatively close. The best result occurs when using the data from 2018M7 to 2019M5 as the training set and the 2019M6 data as the validation set, the testing set AUC reaches 0.7220. This result is lower than using all the original data directly for 5-fold cross-validation, since the 2019M6 data, which is closest to the testing set distribution, is only involved in the validation and not in training.

\textbf{Analysis of Experimental Set 3.}

Experimental Set 3 uses all the data for the 5-fold cross-validation, and the prediction results of the adversarial validation are added as sample weights in the training process, so there is only 1 experiment. The AUC result of adversarial validation is 0.9681, much higher than 0.5, which indicates that the classifier can easily distinguish the training data from the test data, and the two are indeed inconsistent in distribution.

In Experimental Set 3, the final AUC obtained for the validation and testing set are 0.7149 and 0.7202, respectively, which is rather inferior to the benchmark of using the full data directly for the 5-fold cross-validation. This indicates that changing only the sample weights without changing the sample selection does not effectively solve the dataset bias problem.

\textbf{Analysis of Experimental Set 4.}

\begin{figure}[H]
    \centering
    \includegraphics[width=\textwidth]{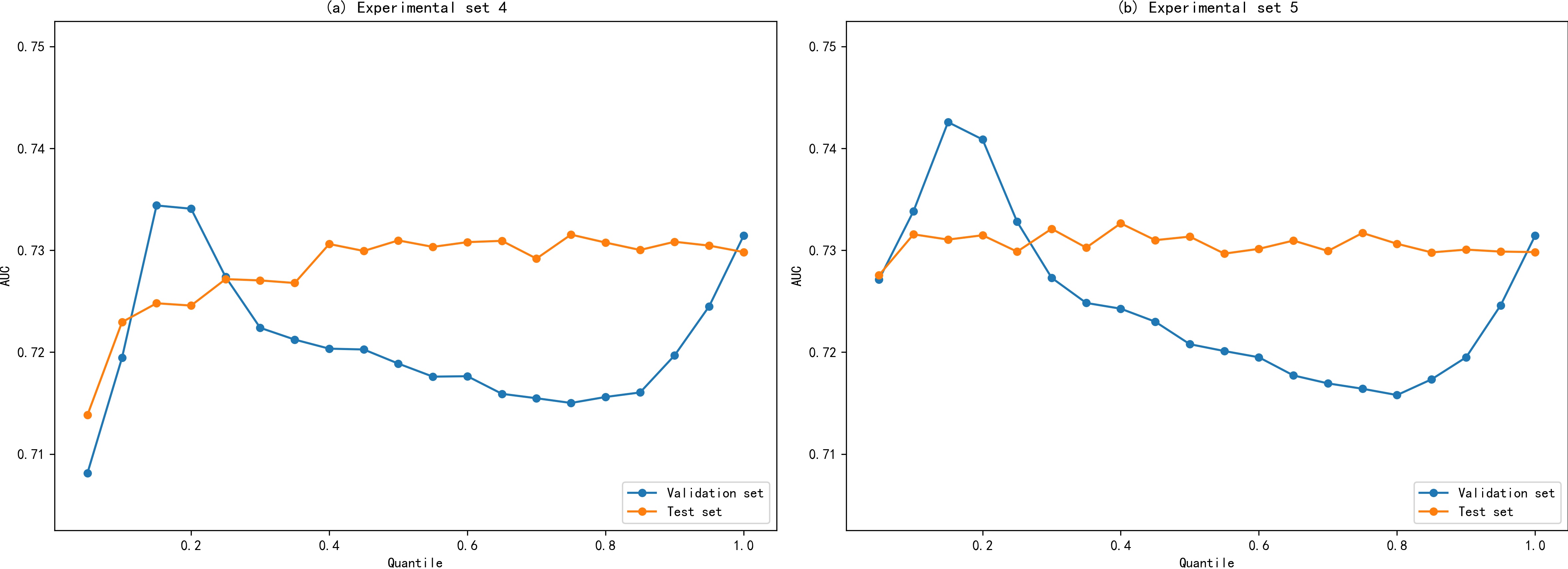}
    \caption{Experimental Set 4 and 5 results.}
    \label{fig: experimental set 4 5}
\end{figure}

According to the results of adversarial validation, Experimental Set 4 selected the data with the top specific quantile for 5-fold cross-validation. Fig \ref{fig: experimental set 4 5} (a) showed the results, with the increase of the probability quantile of adversarial validation, the training data also increased, and the testing set AUC showed a gradual rise at first, and then a relatively stable fluctuation. When the quantile selection is 75\%, the maximum testing set AUC can reach is 0.7315, which is an improvement compared with the previous three experiments. When the quantile is small, the performance of the model is greatly reduced due to the lack of available training data, exposing the drawback that this method fails to make full use of all data.

\textbf{Analysis of Experimental Set 5.}

In Experimental Set 4, the quantile selection has a greater impact on the experimental results. Experimental Set 5 adds the data with the lower quantile ranking to each fold of the training data, but does not participate in the validation phase. This ensures the ability of the model to generalize to the testing set while making full use of all the data. Fig \ref{fig: experimental set 4 5} (b) illustrates the results, the testing set AUC fluctuation of Experimental Set 5 is relatively more stable. When the quantile is chosen to be 40\%, the maximum testing set AUC value can reach 0.7327, which is also the highest score among all experiments.

\begin{figure}[H]
    \centering
    \includegraphics[width=\textwidth]{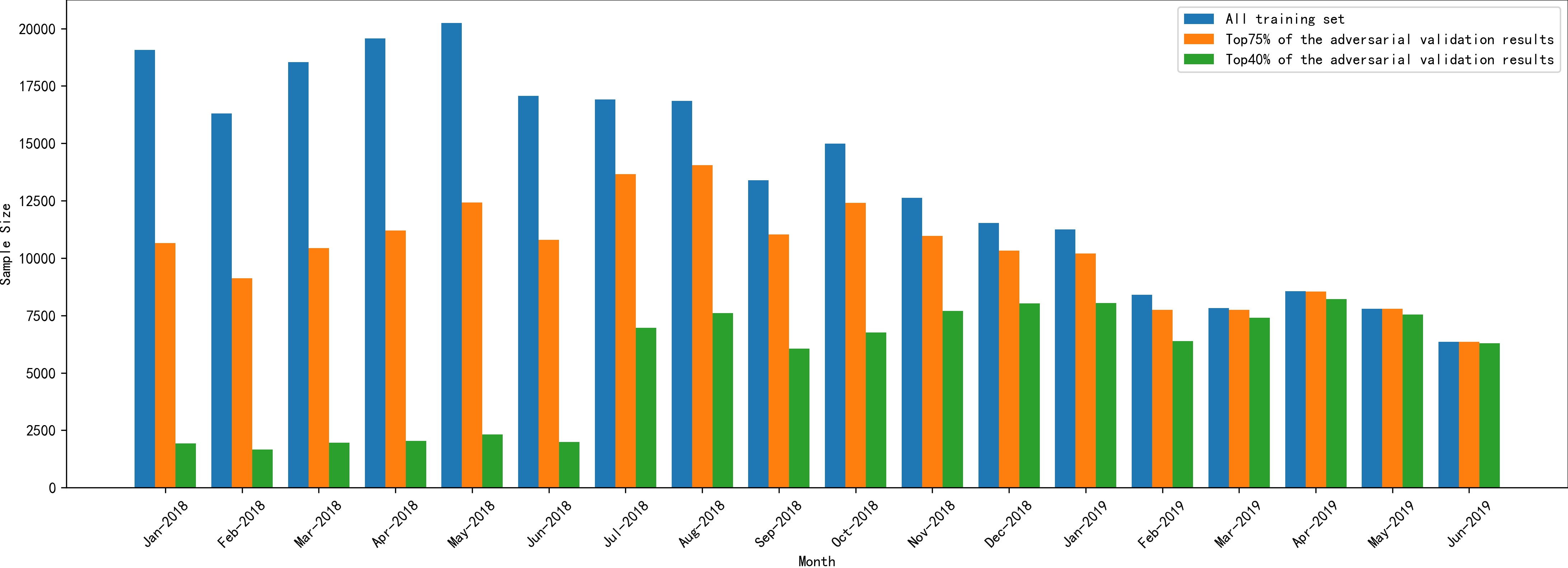}
    \caption{Comparison of sample distribution under different sampling quantiles according to the results of adversarial validation.}
    \label{fig: Comparison of sample distribution}
\end{figure}

In Experimental Sets 4 and 5, the data partitioning quantiles that achieve the optimal accuracy are 75\% and 40\%, respectively. Fig \ref{fig: Comparison of sample distribution} shows the distribution of these data in each month. It can be found that the closer to the testing set date, the higher the percentage of the selected data in the month's original data. This is also in line with the phenomenon that the distribution of data at similar times is more consistent.

\textbf{Comprehensive analysis of all experiments.}

\begin{figure}[H]
    \centering
    \includegraphics[width=8cm]{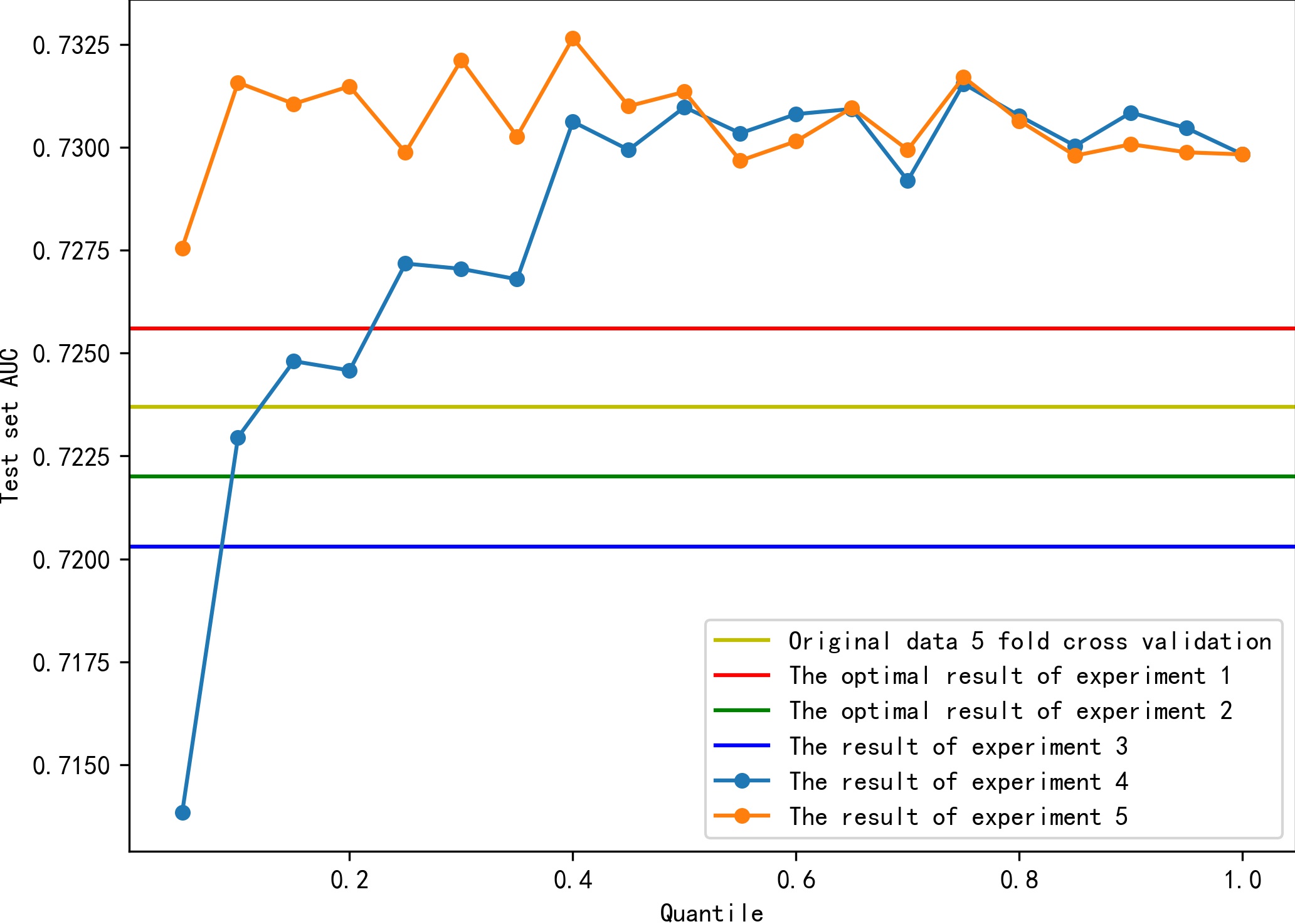}
    \caption{AUC comparison of testing set for all experiments.}
    \label{fig: experiment all}
\end{figure}

Fig \ref{fig: experiment all} shows the comprehensive comparison of the results of all 5 experimental sets, which can be summarized as follows:

\begin{itemize}
    \item The dataset shift problem does exist in credit scoring, and dividing the training and validation sets in different ways will indeed affect the model performance on the testing set.
    \item Compared with other partitioning or data utilization methods, using adversarial validation to select data more consistent with testing set distribution for cross-validation can improve the optimal accuracy. However, attention should be paid to the amount of data selected, too little data may degrade the model performance.
    \item To further improve the model performance, data that are not consistent with the testing set distribution can be added to the training process as well, but not involved in the validation. This will allow the optimal accuracy of the testing set to be further improved, and the results obtained by choosing different quantiles are more stable.
\end{itemize}

\section{Conclusion}
\label{Conclusion}
This paper proposes a method based on adversarial validation to deal with the dataset shift problem in the credit scoring field. Only the training samples whose distribution is consistent with the testing set are used for cross-validation to ensure the generalization performance of the model on the testing data. Meanwhile, to make full use of all data information, the remaining training samples whose distribution is inconsistent with the testing set are added into the training process of each fold of cross-validation, but not involved in the validation.

Experiments on the Lending Club dataset showed that the proposed method is more helpful in improving performance in scenarios where the data distribution of the training set and the testing set are inconsistent, rather than dividing data in chronological order. This work demonstrates the importance of the dataset shift problem in credit scoring. For the sake of performance on new data, it recommends paying more attention to the impact of data distribution on the model effectiveness, rather than just minimizing the classification error.

In the future work plan, more ways to exploit adversarial validation partitioned data can be explored. Transfer learning, which aims to improve the performance of models in different but related target domains, would be a good choice. In addition to credit scoring, the application of adversarial validation can be explored in other data distribution inconsistency scenarios.


\section*{Acknowledgments}
This work was supported by HuaRong RongTong (Beijing) Technology Co., Ltd. We acknowledge HuaRong RongTong (Beijing) for providing us with high-performance machines for computation. We also acknowledge the anonymous reviewers for proposing detailed modification advice to help us improve the quality of this manuscript.

\bibliographystyle{elsarticle-num}
\bibliography{sample}

\end{document}